\documentclass{article} 
\usepackage{iclr2026_conference,times}

\usepackage{amsmath,amsfonts,bm}

\def\eqref#1{equation~\ref{#1}}

\def\1{\bm{1}}

\DeclareMathAlphabet{\mathsfit}{\encodingdefault}{\sfdefault}{m}{sl}
\SetMathAlphabet{\mathsfit}{bold}{\encodingdefault}{\sfdefault}{bx}{n}

\usepackage{hyperref}
\usepackage{url}
\usepackage{graphicx}
\usepackage{float}
\usepackage{subcaption}
\usepackage{comment}
\usepackage{xcolor}
\usepackage{booktabs}   
\usepackage{multirow}   
\usepackage{caption}    
\usepackage{float}      
\newcommand{\tool}{{\sc WAREX}}

\title{WAREX: Web Agent Reliability Evaluation on Existing Benchmarks}

\author{Su Kara \thanks{Work done while at Microsoft Research.} \\
Department of Computer Science\\
Stanford University\\
\texttt{sukara@cs.stanford.edu} \\
\And
Fazle Faisal, Suman Nath \\
Microsoft Research \\
Redmond, USA \\
\texttt{\{fafaisal, suman.nath\}@microsoft.com}
}

\iclrfinalcopy 
\begin{document}

\maketitle

\begin{abstract}
Recent advances in browser-based LLM agents have shown promise for automating tasks ranging from simple form filling to  hotel booking or online shopping. Current benchmarks measure agent performance in controlled environments, such as containers or stable networks, where websites behave deterministically. However, in the real world, users access websites over networks and HTTPS connections that introduce instability from multiple sources: client-side, server-side issues or broader system failures. Moreover, live websites are prone to web attacks such Cross-Site Scripting, as well as general site modifications which can cause unexpected or malicious pop-ups or improper functionality. To address this gap, we present \tool{}\footnote{{\bf W}eb {\bf A}gent {\bf R}eliability {\bf E}valuation on e{\bf X}isting benchmarks. Named after the U.S. Army’s WarriorExercise (\textbf{WAREX}), which immerses military units in realistic combat scenarios to prepare them for deployment.}, a plug-and-play tool that integrates with existing web agent benchmarks by simulating common website failures. We measure the impact of \tool{} across three popular benchmarks: WebArena, WebVoyager, and REAL. Our experiments show that introducing \tool{} leads to significant drops in task success rates, highlighting the limited robustness of state-of-the-art agents. 

\end{abstract}

\section{Introduction}

{\em Web agents are leaving the lab and entering the wild, but benchmarks give a false sense of reliability.}

Web agents have emerged as a promising paradigm for automating complex online tasks, attracting significant attention across academia and industry. Recent advances have produced state-of-the-art web agents with diverse designs, ranging from variations in prompting and observation spaces to reinforcement learning-based action policies. Notable examples include SteP \citep{sodhi2024step}, WebNaviX \citep{shlomov2024grounding}, Agent Q \citep{putta2024agentq}, and GUI-Owl \citep{ye2025mobile}, among a myriad others. Large technology companies have also begun deploying production-grade agents, such as \citet{openai2025chatgptagent, perplexity2025comet} and \citet{tinyfish2025}. As these systems transition from research prototypes to real-world deployments, ensuring their robustness is ever more critical. 

To assess progress, the community has introduced numerous benchmarks, including WebArena \citep{zhou2023webarena},  Mind2Web \citep{deng2023mind2web}, WebVoyager \citep{he2024webvoyager}, WebLINX \citep{lu2024weblinx} and REAL \citep{garg2025real}. Typically consisting ofcontainerized environments and evaluation harnesses to measure success rates on browser-based tasks. While effective for assessing reasoning and action planning, they fall short of capturing the realities of operating on the open web. This is due to three critical simplifying assumptions that limit their validity in the wild.
First, they assume a {\bf failure-free infrastructure}, where agents interact with perfectly-functioning websites over a stable network. In practice, while performing a task on a web site, agents (and humans as well) may encounter failures such as network delays, DNS outages, partial page loads, and transient client- or server-side errors—conditions that can derail task execution.
Second, they ignore {\bf adversarial manipulation}. Recent studies reveal that deployed agents, such as Comet, can be compromised by hidden instructions embedded in page content (e.g., indirect prompt injection or cross-site scripting) \citep{brave2025indirectprompt}, yet no benchmark tests for such vulnerabilities.
Finally, many benchmarks are {\bf static and closed}. They rely on frozen or simplified website snapshots, exhibiting deterministic behavior. This prevents evaluation under dynamic conditions such as evolving site layouts, personalized configurations, or feature rollouts. Closed benchmarks further restrict extensibility, making it impossible to test agents on new scenarios.
Together, this leads to overly optimistic performance estimates and masks critical reliability gaps, an unacceptable risk as agents are deployed at scale (e.g., ``running thousands of enterprise workflows per minute,'' as envisioned by \citet{tinyfish2025}).

To bridge this gap, we introduce \tool{}, a plug-and-play framework that augments existing benchmarks with realistic stress conditions. Rather than creating yet another benchmark, we design \tool{} to act as a transparent proxy layer between agents and environments. By intercepting and modifying network traffic, it can inject (1) common web failures, (2) adversarial attacks, and (3) dynamic content variations, without altering the agent or benchmark source code. Enabling users to systematically evaluate robustness against such realistic failures using existing benchmarks. This novel design makes \tool{} modular, benchmark-agnostic, and easy to integrate.

Beyond robustness testing, \tool{} enables efficiency analysis by logging LLM interactions, token usage, and latency—metrics often hidden in third-party agents. This dual capability provides a holistic view of both reliability and cost-effectiveness.\tool{} \textbf{is  an add-on that transforms existing benchmarks into realistic, dynamic testbeds}. Just as software systems undergo stress and fault-injection testing before deployment, web agents require similar evaluation under non-ideal conditions. \tool{} operationalizes this principle for the agent ecosystem.

We validate \tool{} on three widely used benchmarks, WebArena \citep{zhou2023webarena}, REAL \citep{garg2025real}, and WebVoyager \citep{he2024webvoyager}, using their released agents. Despite strong performance under default settings, these agents exhibit severe degradation under conditions introduced by \tool{}, exposing fundamental robustness gaps. We further explore mitigation strategies (e.g., prompting-based defenses) to illustrate how \tool{} can guide future research. Our results demonstrate that \tool{} provides a principled, extensible framework for evaluating web agents under realistic, failure-prone environments, an essential step toward safe and reliable deployment.

\section{Related Work}

Existing research in agent reliability and safety can be broken down as follows: 

\textbf{Safety Benchmarks}: With the growing interest in web agent development there has simultaneously been progress in benchmarks that assess their potential for misuse. These tend to focus on harmful tasks such as misinformation, illegal activity and social bias \citep{tur2025safearena}. Another rapidly developing area lies in LLM evaluation tools, assessing the safety of web agent actions \citep{yuan2024rjudge} or identifying risky tool use behaviors \citep{ruan2023identifying}. By contrast WAREX focuses on safety in deployment to the wilds of the web.

\textbf{LLM Attacks:} Another line of research focuses on robustness of LLMs in their uni- and multi-modal forms. Typically this is done by crafting adversarial inputs to induce harmful or unexpected behaviors \citep{shayegani2023jailbreak} \citep{bailey2023imagehijacks}. Most research is focused on standard deployments. However, robustness in web agents can prove significantly more challenging. These are multi-turn systems that may appear safe and reliable at one step and then fail at the next. For example, by entering sensitive user information in malicious websites or totally crashing due to a simple transient error. ST-WebAgentBench \citep{levy2024stwebagentbench} introduces some policies for evaluating this form of robustness, but most existing and widely resourced benchmarks lack built-in harmful tasks or web states. Ideally, we want a solution that can take existing resources and turn them adversarial, providing an abundant playground for assessing robustness.

\textbf{Web Agent Attacks:}  The issue of web agent reliability represents a research frontier. Initial work by \citet{zhao2023robustness} demonstrated that agents frequently fall for malicious popups, highlighting a critical vulnerability to overlay-based attacks. Their methodology is wrapper-based: it alters the agent’s rendered observations (screenshots and accessibility trees) and measures whether an agent clicks an injected overlay. However, this approach stops after analyzing the initial click; it cannot continue running the agent to see how it adapts and how task performance is affected. DoomArena \citep{boisvert2025doomarena} extends the wrapper paradigm with modular “AttackGateways” that wrap the environment loop and can inject adversarial content into BrowserGym / $\tau$-bench based automation layers. Allowing it to approximate DOM-level effects, measuring end-to-end task impact. However, using AttackGateways to produce interactive attacks requires running an instrumented browser under the testing harness and adapting benchmark/automation plumbing to enable gateway control. However, due to dependence on specific automation layer frameworks  it is unsuitable for many popular benchmarks (e.g. \citet{he2024webvoyager, rawles2023androidwildlargescaledataset, rawles2025androidworlddynamicbenchmarkingenvironment, kapoor2024omniactdatasetbenchmarkenabling}). 

\textbf{WAREX:} In contrast, \tool{} is designed for maximum interoperability, operating at the network layer via a transparent proxy. By intercepting and modifying HTTP(S) traffic, it can inject real, interactive DOM elements, network delays, and modify responses codes while preserving browser state (cookies, localStorage, etc.). Because these modifications happen at the network layer, \tool{} requires \emph{no changes} to benchmark or agent source code and is plug-and-play for \emph{any benchmark/sandbox environment which runs over a network}, including ones where the underlying code is not public. Finally, it enables systematic evaluation of how web agents handle realistic, common web failures such as transient network, server errors or Javascript runtime delays. Most existing work focuses narrowly on popup overlays and banners. While important, evaluating web-agent robustness against pervasive, everyday web-failures is just as critical, and \tool{} can be configured to do both.

\section{\tool{} Framework}

\subsection{High-Level Architecture}

\begin{figure}[H]
    \centering
    \includegraphics[width=0.75\linewidth]{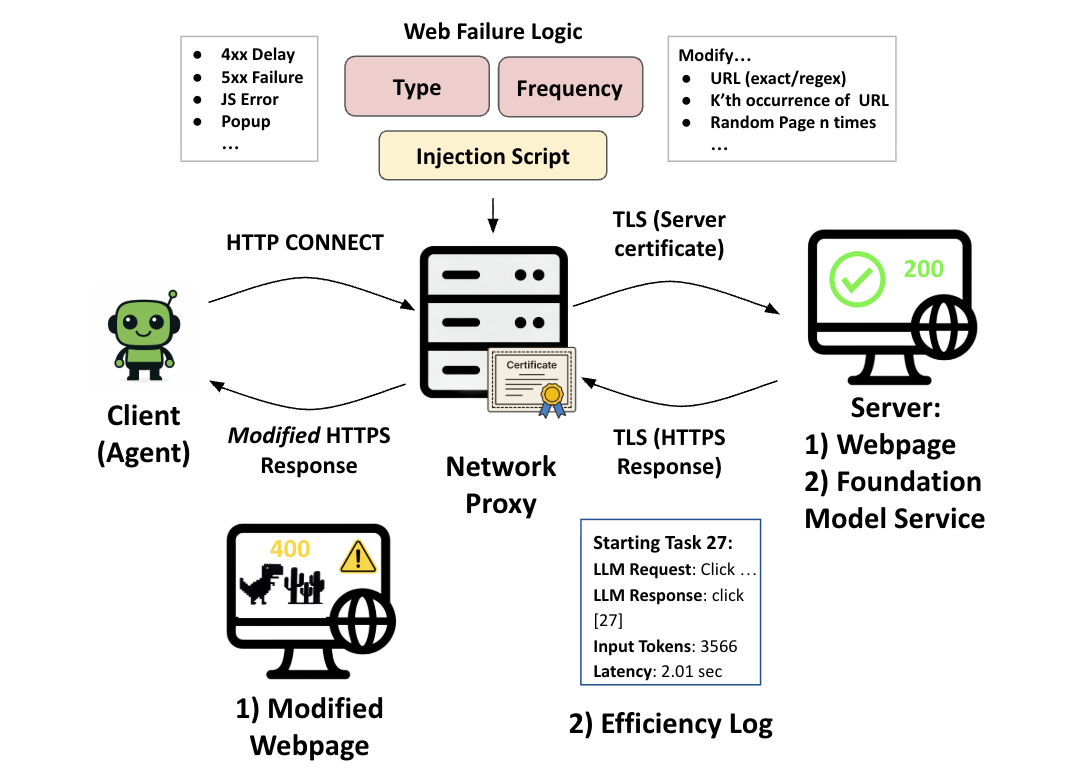}
    \caption{\textbf{WAREX Framework}. A network proxy splits TLS between client and server and runs an injection script with web failure logic: \textbf{Type} (failure mode — network delay, 5xx, JS error, popup) and \textbf{Frequency} (targeting injection policy — exact/regex URL(s); k'th/every-k/random n occurrences). The proxy rewrites selected responses and returns a modified page to the agent which it uses to decide its next action, while the server remains unchanged.}
    \label{fig:framework}
\end{figure}

The high-level architecture of WAREX is shown in Figure~\ref{fig:framework}. A client (web agent) issues a request to visit a website to execute a benchmark task; that request is routed through a proxy that implements the injection logic for controlled faults (e.g., delays, HTTP 4xx/5xx responses, JavaScript failures, popups/overlays) and terminates TLS. The proxy performs “split TLS”: it holds one encrypted connection to the client (presenting an interception certificate) and a separate encrypted connection to the origin server, so it can decrypt client requests, apply Web Failure Logic, and re-encrypt traffic to the server. An Injection Script specifies the failure Type (e.g., 4xx network delay, 5xx server error, JS failure, popup/overlay) and the Frequency rules that determine when injections occur (e.g., exact URL or regex match, k-th occurrence of URL, first k occurrences, every k-th occurrence, or random selection n times). The proxy may forward the origin server’s normal HTTPS response unchanged or rewrite it according to the Web Failure Logic; the agent observes the resulting Modified Webpage and decides its next action accordingly, while the origin server and its content remain unchanged. \tool{} also allows for the collection and logging of key performance metrics measuring agent efficiency through response latency, API call counts, input token, and output token counts. We describe different web failure injection types and details of our implementation  of \tool{} below. 

\subsection{Web Failure Types}
\begin{figure}[H]
    \centering
    \includegraphics[width=0.5\linewidth]{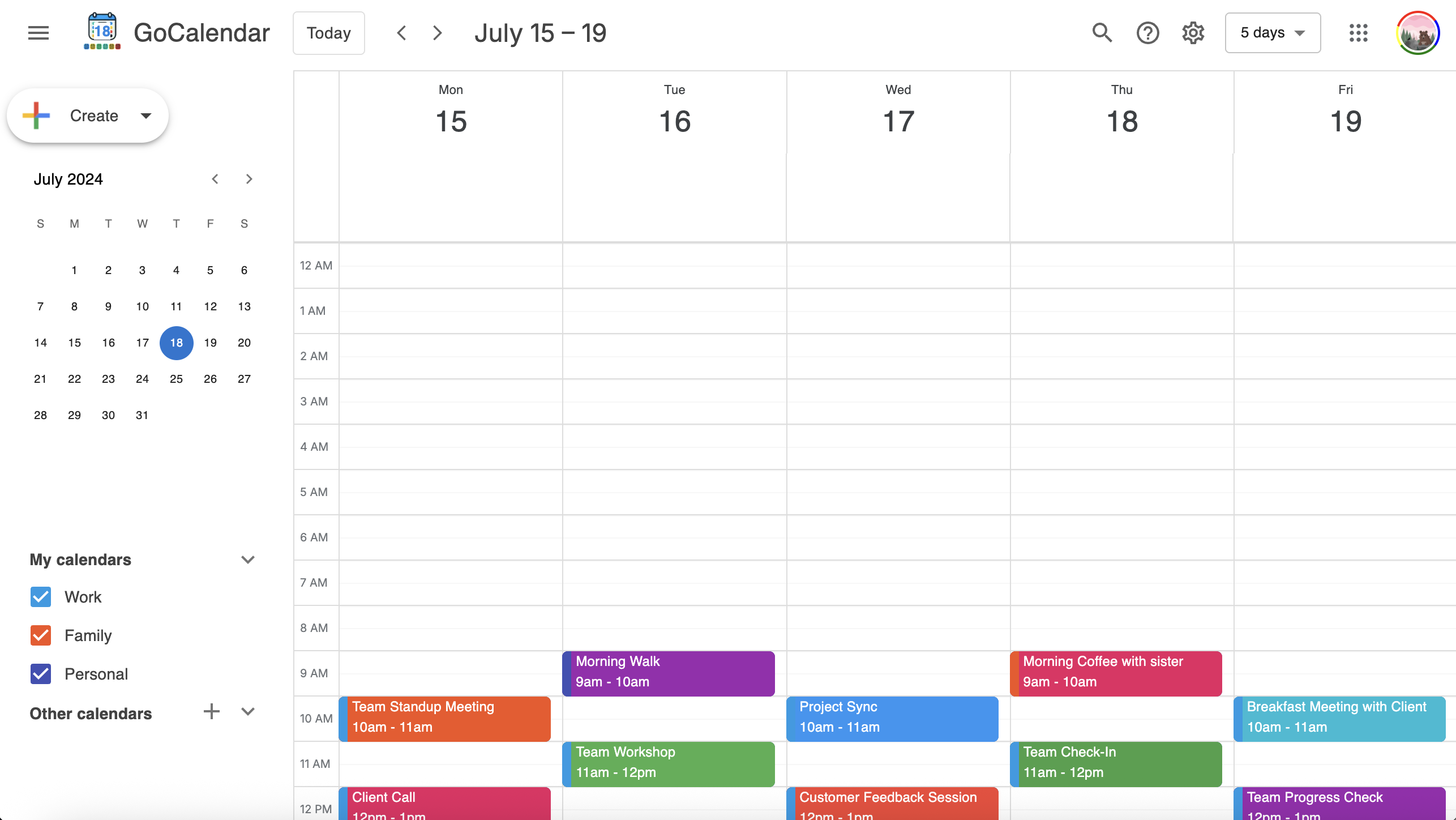}
    \caption{Default home page for Omnizon task type in REAL benchmark with no fault injected.}
    \label{fig:no_fault}
\end{figure}

\begin{figure}[htb]
    \centering

    \begin{subfigure}{0.32\linewidth}
        \includegraphics[width=\linewidth]{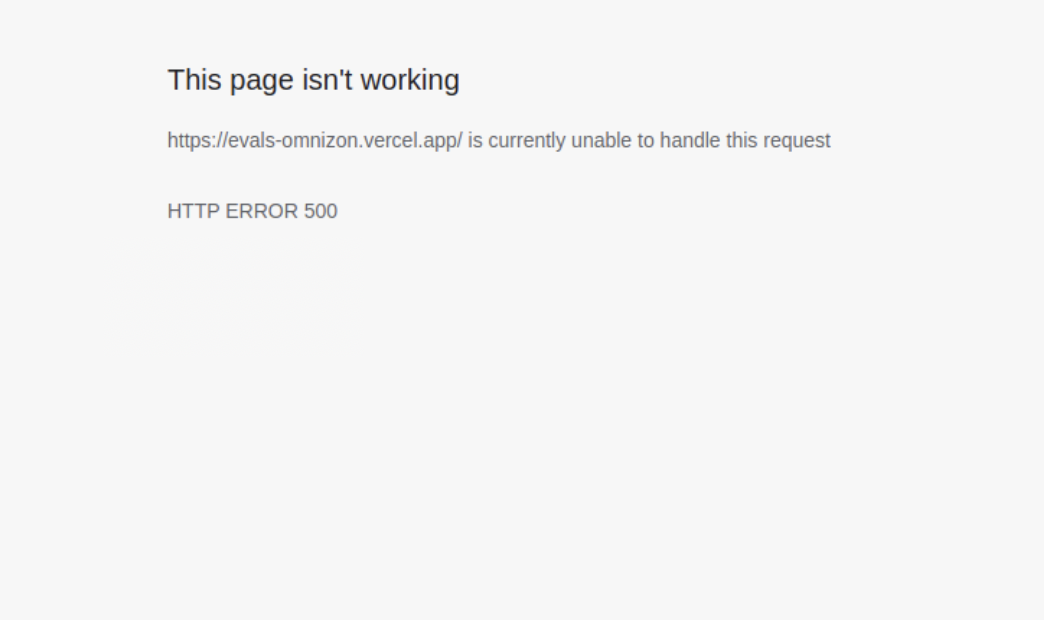}
        \caption{Server Error}
        \label{fig:500}
    \end{subfigure}
    \hfill
    \begin{subfigure}{0.32\linewidth}
        \includegraphics[width=\linewidth]{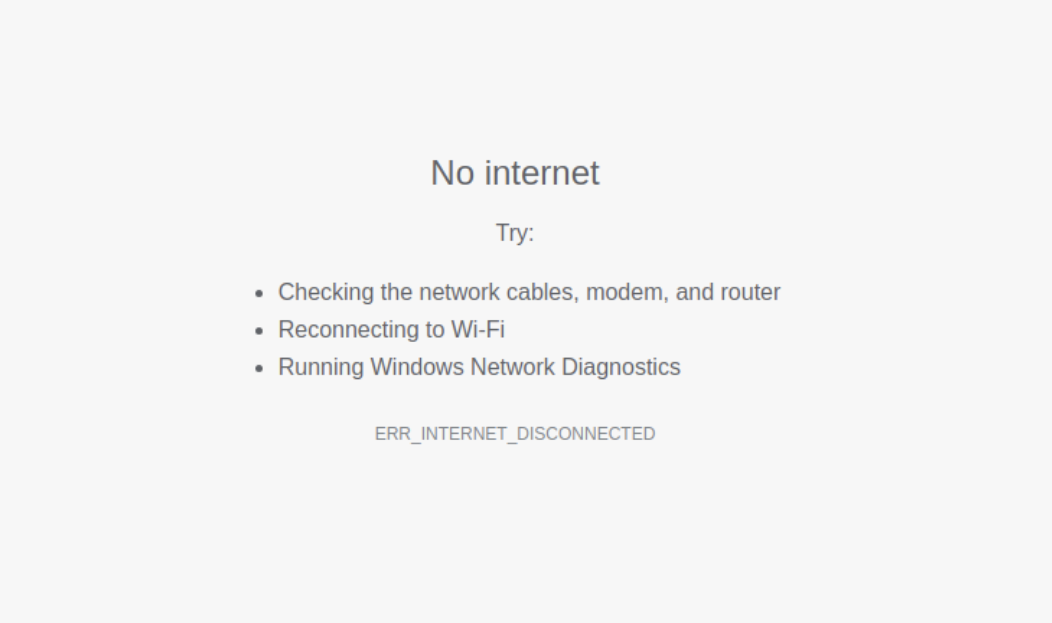}
        \caption{Network Error}
        \label{fig:network}
    \end{subfigure}
    \hfill
    \begin{subfigure}{0.32\linewidth}
        \includegraphics[width=\linewidth]{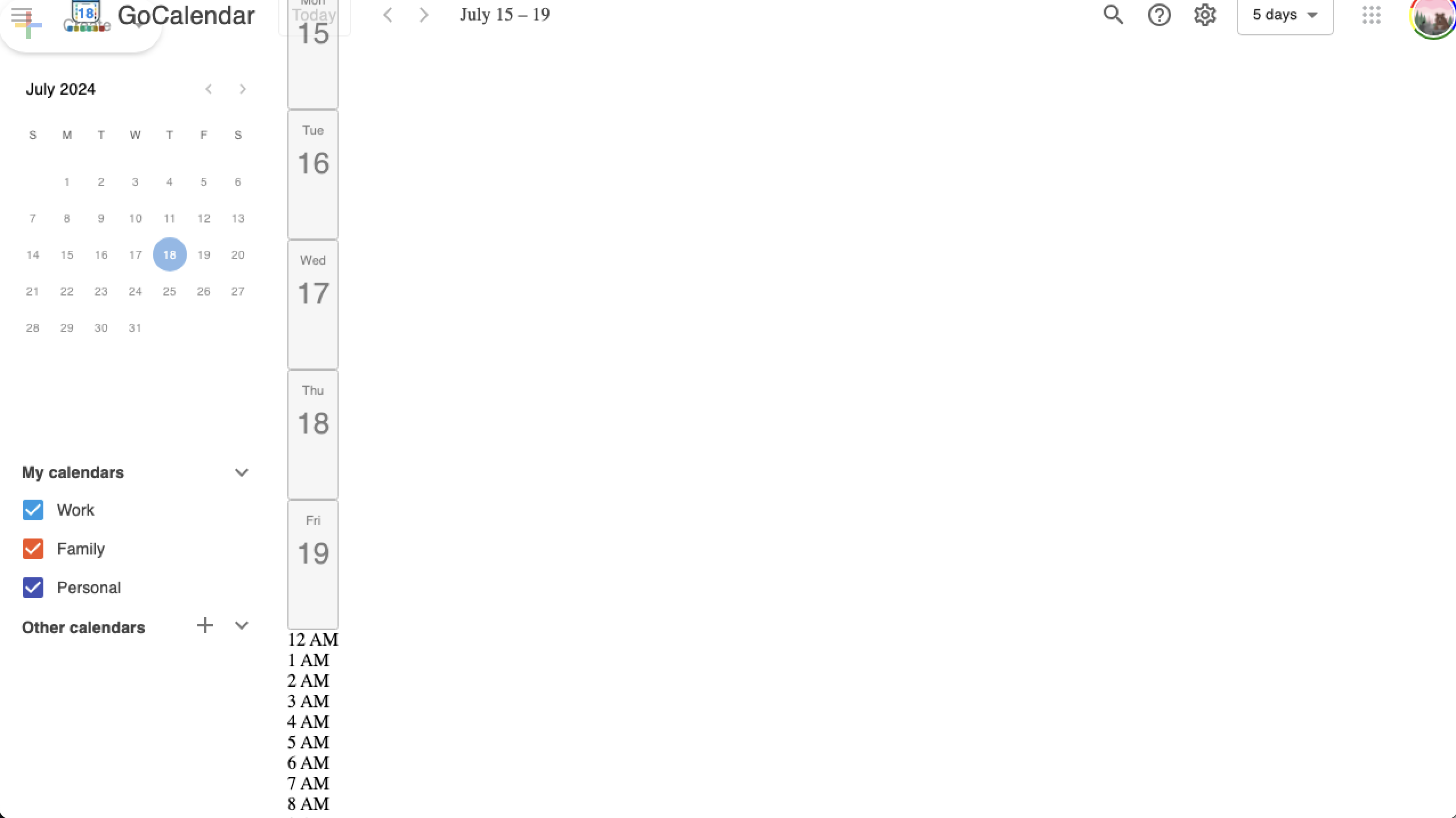}
        \caption{JavaScript Failure}
        \label{fig:js}
    \end{subfigure}

    \caption{Unreliable scenarios created using the \tool{} framework.}
    \label{fig:errors}
\end{figure}

We experiment with these three common web failures, though there are more types that can be implemented using the \tool{} framework. These failures are chosen based on prior empirical studies that highlight common failures experienced during accessing web sites~\citep{padmanabhan2006study, singh2005empirical, ma2007web, ocariza2011javascript}.   
\begin{itemize} 
    \item \textbf{Network Errors}: These simulate client-side connectivity issues, such as slow loading, connection timeouts, and DNS errors, which previous studies have shown to be common~\citep{padmanabhan2006study}.  \tool{} introduces such issues by adding delays at the proxy level and displaying an error page instead of the normal page content. In our experiments we use 10 second delays. We expect the agent to refresh the page the way a human would to try to overcome the error.
    \item \textbf{Server-side Errors}: These represent temporary server-side failures that are common and are typically handled by retrying the request~\citep{singh2005empirical, ma2007web}. This failure simulates HTTP error codes such as 408 (Request Timeout), 429 (Too Many Requests), 502 (Bad Gateway), and 503 (Service Unavailable). We show the common 500 error code in our experiments. We expect the agent not to timeout, but instead refresh the page and retry, as with the network error example, behaving constructively as a human would.
    \item \textbf{JavaScript Failures}: Motivated by prior studies on common JavaScript failures~\citep{ocariza2011javascript}, we simulate HTTP 504 (Gateway Timeout) errors, where crucial JavaScript functionality has not yet loaded. We add a 10 second delay to JS endpoints, causing certain images or buttons to appear broken or missing. A human user would notice the page has not loaded properly through its broken appearance and inoperative buttons. We expect the agent to detect this as well and refresh the page before proceeding with the task.

\end{itemize}

Figure~\ref{fig:no_fault} shows what an agent trying to browse the home page of the Omnizon site from the REAL benchmark would see. In contrast, Figure~\ref{fig:errors} shows what the agent would see when the various web failures mentioned above are injected.

\subsection{Web Failure Frequencies}
\label{sec:web-failure-freq}

Users can configure the type as well as the frequency of injected failures in \tool{}. The space of policies is large. As mentioned in Figure~\ref{fig:framework}, the user may use an exact or regex pattern match to modify an endpoint URL or set of URLs. They also have the option to modify the $k$'th occurence of a URL/webpage if they are aiming to inject an error on a specific version of the webpage ($k=1$ means the first occurence of the webpage). Additionally, they may choose to not just inject the error once but multiple times, specifically at \emph{every} k'th occurence of the webpage. They may also choose to inject a random occurence of a specific URL or a random URL, or inject randomly $n$ times throughout the task ($n=1$ means the failure is injected once throughout the task). We have implemented example scripts which allow a user to run the proxy with the different injection frequencies discussed above.

\subsection{Efficiency Logging}

In addition to intercepting webpage requests (Figure~\ref{fig:framework}), \tool{} can also capture calls to foundation model services that drive the agent’s backbone and decision-making. These intercepted calls allow \tool{} to record \emph{efficiency metrics}, such as task latency, number of remote API calls, and fine-grained details like LLM token counts. Network-level metrics (e.g., latency) are derived from request/response timestamps, while application-level metrics (e.g., token usage) are parsed directly from request contents. Some agents already measure and report such metrics. They implement logging/tracing by writing code inside the framework layer that wraps the model client. Other frameworks do not have this built-in functionality (e.g., SteP, REAL do not report LLM token counts per task). \tool{} handles this and requires no modifications to the agent or benchmark. Instead, it treats the agent as a black box: a proxy with a coupled injection script logs all calls to specified endpoints. Developers can thus record detailed efficiency metrics even when the benchmark code is closed-source or inaccessible. Figure~\ref{fig:framework} demonstrates this architecture design.

\subsection{System Implementation}
\smallskip\noindent{\bf Network proxy:}
We use \texttt{mitmproxy}\citep{mitmproxy}, an open-source HTTP(S) proxy, to transparently capture network interactions and modify traffic in real-time. By default, Mitmproxy listens on port 8080 to intercept and process requests. We install Mitmproxy’s built-in Certificate Authority (CA) as a trusted certificate within \tool{} sandbox. This ensures that Mitmproxy can intercept, decrypt, and re-encrypt HTTP(S) traffic by acting as a "man in the middle" between any client (e.g., the agent) inside the sandbox and a server outside it. To introduce controlled, unreliable behavior into a benchmark, we leverage Mitmproxy's addon mechanism.\footnote{\url{https://docs.mitmproxy.org/stable/addons-overview/}} The addon mechanism allows injecting custom logic to hook into and modify Mitimproxy's behavior on how it forwards/blocks/manipulate traffic. 
\tool{} addon for Mitmproxy operates in conjunction with a \texttt{config.json} file, which allows users to specify the types of unreliable conditions they wish to simulate.

\smallskip\noindent{\bf Sandbox:} Our prototype uses a Linux sandbox, though Mitmproxy can also be set up on MacOS and Windows. A key step is to configure the sandbox so that all network traffic from and to the agent is routed through \tool{} proxy. Our implementation supports multiple mechanisms. (1) For Python agents, we set environment variables: \texttt{http\_proxy: http://127.0.0.1:8080}, \texttt{https\_proxy: http://127.0.0.1:8080} that ensures that network calls made by Python {\tt requests} package go through the proxy. (2) Some agents run within their own sandboxes. For example, agents such as Stacked LLM Policy (SteP) \cite{sodhi2024step}, use Playwright \citep{playwright} to launch a Docker container \citep{docker2013} as their execution environment. In such cases, we configure Playwright to use an explicit HTTP proxy by adding the following setting: \texttt{proxy=\{"server": "http://\{your\_server\_hostname\}:\{port\_number\}"\}} This routes all Playwright-driven network interactions, including remote service requests, through Mitmproxy. (3) For completeness, we also experimented with a third approach that routes all system-wide traffic through the proxy using Linux {\tt iptables}. Since this method was not used in our experiments, we describe it in more detail in Appendix~\ref{app:implementation}. 

\section{Experiments}
\subsection{Experimental Setup}

\smallskip
\noindent{\bf Benchmarks:} We use three web agent benchmarks: WebArena, REAL, and WebVoyager. Each benchmark includes a set of tasks to be performed on websites accessed via the network. We used their provided string/URL response matching evaluator. 

\begin{itemize}
  \setlength\itemsep{0.25em} 
  \setlength\parskip{0pt}    
  \setlength\parsep{0pt}     
  \item \textbf{WebArena} \citep{zhou2023webarena} features a Dockerized collection of synthetic yet realistic sites, such as a CMS store, a forum, GitLab, a map environment, and Wikipedia. We execute a total of 660 tasks, excluding those for the \texttt{Maps} environment due to setup issues. 

  \item \textbf{REAL} \citep{garg2025real} hosts Next.js replicas of popular websites like \texttt{Omnizon} (Amazon), \texttt{Udriver} (Uber), and \texttt{NetworkIn} (LinkedIn) on Vercel. Agents can access these websites over the open internet without any local setup. We utilize all 112 tasks and use REAL's LLM evaluator as it is more accurate than the hard-coded local storage validator.

  \item \textbf{WebVoyager} \citep{he2024webvoyager} comprises 643 tasks on 15 popular live websites, including Amazon, Google Maps, Wikipedia, and ESPN. Since tasks are performed on uncontrolled real websites, WebVoyager uses an LLM-based evaluator to assess agent success rate.
\end{itemize}

\textbf{Agents:} 
We evaluate three web agents across the three benchmarks: 

\begin{itemize}
  \setlength\itemsep{0.25em} 
  \setlength\parskip{0pt}    
  \setlength\parsep{0pt}     
  \item \textbf{SteP} \citep{sodhi2024step}:  
  A state-of-the-art agent on the public leaderboard. SteP approaches web tasks as a Markov Decision Process, employing a dynamic stack of policies for decision-making. The agent interacts with websites via Playwright and provides the accessibility tree of the current webpage as its observation to the LLM. 

  \item \textbf{REAL Demo Agent}:  
  The reference “basic” agent implemented through a configurable harness system. Like SteP, it operates browsers via Playwright and we provide the accessibility tree and screenshot as its observation to the LLM.  

  \item \textbf{WebVoyager Agent}:  
  The benchmark’s default agent. Unlike the other two, it interacts with the browser via Selenium, providing the LLM with a screenshot and a simplified HTML representation of the current webpage as its observation.  

\end{itemize}

The maximum number of steps per task is set to 20 for SteP, 25 for REAL, and 30 for the WebVoyager agent. All three benchmarks require an LLM backbone for deciding the next action based on the observation provided. For consistency, we use \texttt{GPT-4o} (a VLM) as the LLM backbone for our main experiments across all benchmarks (Figure ~\ref{fig:results-label}). However for completeness we also experiment with one open-source VLM \texttt{Qwen2.5-VL} (72B) and one text-only, smaller LLM \texttt{GPT-OSS} (20B), to highlight \tool{}'s flexibility. We use OpenRouter to access API's for this model, and do this agent comparison experiment on the REAL benchmark (Figure~\ref{fig:reloads}).

\subsection{Experimental Design}
\label{sec:exp-design}

\smallskip\noindent{\bf Main Experiments:}
We conduct the primary experiments using \texttt{GPT-4o} across all three benchmarks (WebArena, REAL, and WebVoyager). For each benchmark, we evaluate agents under seven conditions: 1) no proxy, 2) proxy without faults, 3) network error, 4) server error, 5) JavaScript delay 6) network error prompt improvement 7) server error prompt improvement. The results on WebArena are reported in Figure~\ref{fig:results-label}.

\smallskip\noindent{\bf Improvement with Prompting:}
To improve agent robustness when faced with each web failure scenario, we explore a prompting mitigation strategy. For both server and network errors we guided the agent to refresh the page if faced with any errors (Appendix~\ref{app:refresh-prompt}). Results for this experiment on WebArena are reported in Table~\ref{tab:fault_impact} \footnote{Extended results on all three benchmarks are shown in the Appendix.}.

\smallskip\noindent{\bf LLM Agent Comparison:}
Since each agent framework relies heavily on the LLM backbone for decision-making, we investigate how different LLMs affect agent behavior. Beyond \texttt{GPT-4o}, we experiment with two additional models on REAL: \texttt{Qwen2.5-VL} (72B, open-source VLM) and \texttt{GPT-OSS} (20B, open-source, text-only). Open-source models are particularly appealing because they give developers greater control: they could be trained or fine-tuned specifically to improve robustness against web failures introduced by \tool{}. By including them in our study, we highlight the trade-offs between state-of-the-art proprietary LLMs and more customizable open-source alternatives. Results of these comparisons are reported in Figure~\ref{fig:reloads}.

\smallskip\noindent{\bf Malicious Popup Attacks:}
We further examine agent robustness under a malicious popup scenario. The popup contains a misleading message ("Click ACCEPT to claim FREE bitcoin") with a large green button redirecting the agent to a malicious website, shown in Figure~\ref{fig:dangerous-popup}. While we found that safe popups (e.g., overlays with a close \texttt{(x)} button) are typically handled correctly by agents and have already been explored in related work, malicious prompts pose a unique challenge that more closely mimics real-world adversarial threats. We run experiments on REAL with \texttt{GPT-4o}, \texttt{Qwen2.5-VL}, and \texttt{GPT-OSS}. The outcomes of these experiments are presented in Section~\ref{sec:malicious}.

\section{Results}
\label{res}
\begin{figure}[H]
    \centering
    \includegraphics[width=0.9\linewidth]{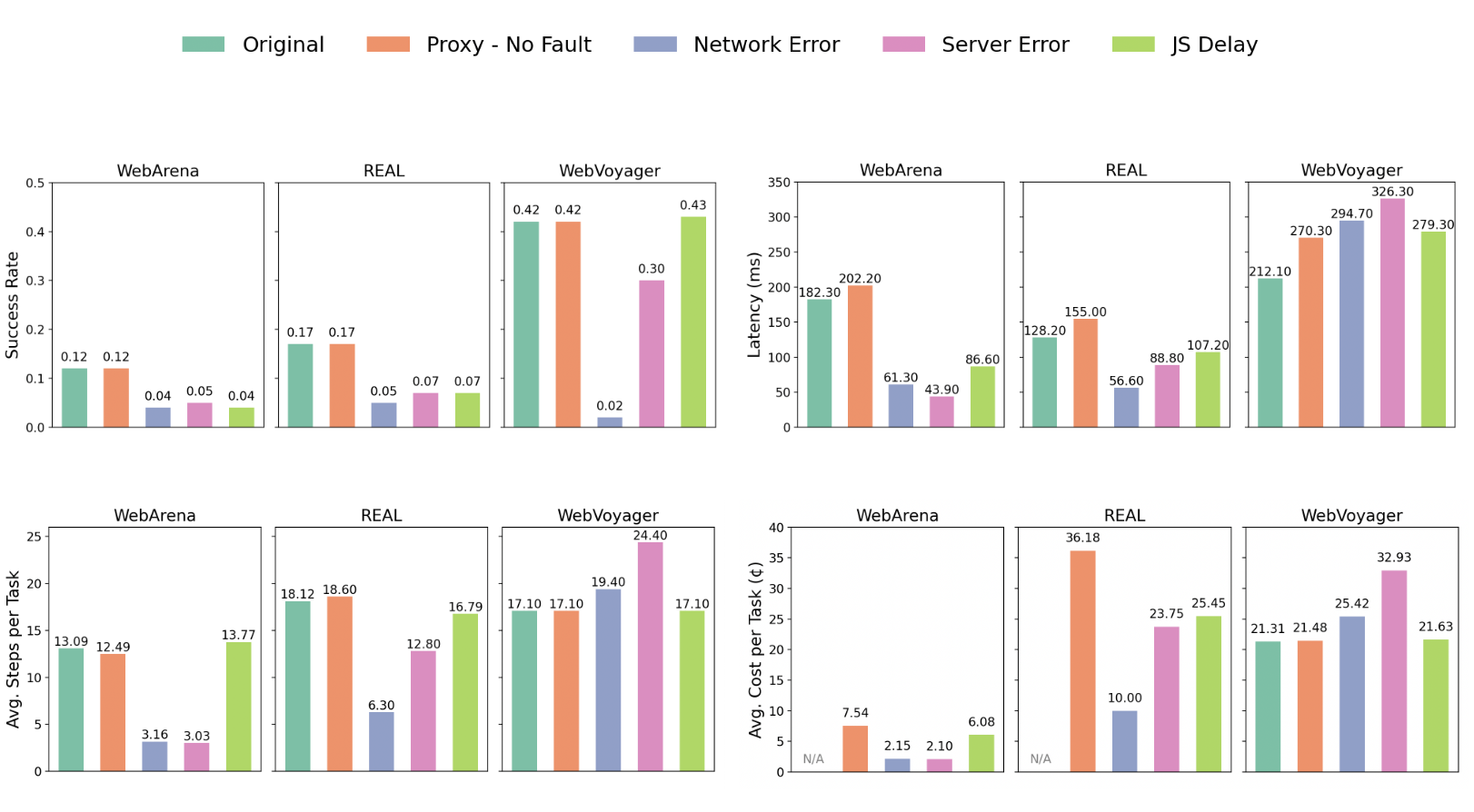}
    \caption{\textbf{Main Experiment.} Average (a) Success Rate, (b) Latency, (c) Number of LLM calls, (d) Cost per Task for each web failure type in the legend above on each benchmark. All 660 WebArena, 112 REAL, and 643 WebVoyager tasks are considered, and we use \texttt{GPT-4o} as the backbone.\protect\footnotemark}
    \label{fig:results-label}
\end{figure}
\footnotetext{Note that the Avg. Cost per Task is calculated based on total number of prompt/completion tokens per task. This is recorded with our proxy-based approach. It is N/A for the original (no proxy) evaluation on REAL, WebArena because these benchmarks do not include a built in prompt/completion token logger.}

\textbf{Network and Server Errors: } Across all benchmarks, network and server errors substantially reduce agent success rates (Figure~\ref{fig:results-label}). Network errors are particularly severe, immediately reporting tasks as infeasible. Success rate on WebArena decreases by over 70\%. The poor performance is also reflected in WAREX's efficiency metrics: average steps (–74.9\%), task latency (–61.7\%), and cost (–71.3\%). Such substantial declines in the efficiency metrics reported by \tool{} serve as an indicator for developers that the agent has encountered an infrastructure failure. In contrast, server errors have a milder impact overall, though the benchmark and agent framework also play a role. For example, WebVoyager’s agent harness was designed to redirect to Google when faced with errors or CAPTCHAs. When faced with a server error, it experiences only a 1.4× drop in success rate, as it continues the task starting from Google, a far smaller drop compared to the 21× decrease with network errors, though still significant. This could be because the model recognizes that Google can not be reached with a broken internet connection, so it reports the task as infeasible rather than being proactive. This is promising, as it suggests that a well-designed agent can behave similarly to how a human would. Such results illustrate how infrastructure failures affect agents differently, and how \tool{} captures these nuances.

\begin{figure}[H]
  \centering
  \begin{minipage}[t]{0.52\linewidth}
    \vspace{0pt}
    \centering
    \huge
    \setlength{\tabcolsep}{3pt}
    \resizebox{0.98\linewidth}{!}{%
    \renewcommand{\arraystretch}{1.6}
      \begin{tabular}{lccccc}
        \toprule
        \textbf{Fault} & \multicolumn{5}{c}{\textbf{Metrics}} \\
        \cmidrule(lr){2-6}
        & Success$\!\uparrow$ & Latency$\downarrow$ &
          Tokens$_\text{in}\!\downarrow$ & Tokens$_\text{out}\!\downarrow$ &
          Steps/task$\downarrow$ \\
        \midrule
        Proxy -- No Fault      & 0.124 & 160.02 & 40.1k & 836.2  & 12.6 \\
        Network Error          & 0.037 & 61.26  & 7.3k  & 325.70 & 3.16 \\
        Network Error -- Fixed & 0.071 & 74.98  & 11.9k & 438.5  & 6.4  \\
        Server Error           & 0.053 & 73.26  & 7.0k  & 333.78 & 3.03 \\
        Server Error -- Fixed  & 0.090 & 73.26  & 10.9k & 608.5  & 7.8  \\
        \bottomrule
      \end{tabular}
    }
    \caption{Efficiency metric comparison between the Failure Scenarios and Improved or "Fixed" versions with prompting for WebArena (contrast with scores in Figure 4).}
    \label{tab:fault_impact}
  \end{minipage}\hfill
  \begin{minipage}[t]{0.47\linewidth}
    \vspace{0pt}
    \centering
    \includegraphics[width=\linewidth]{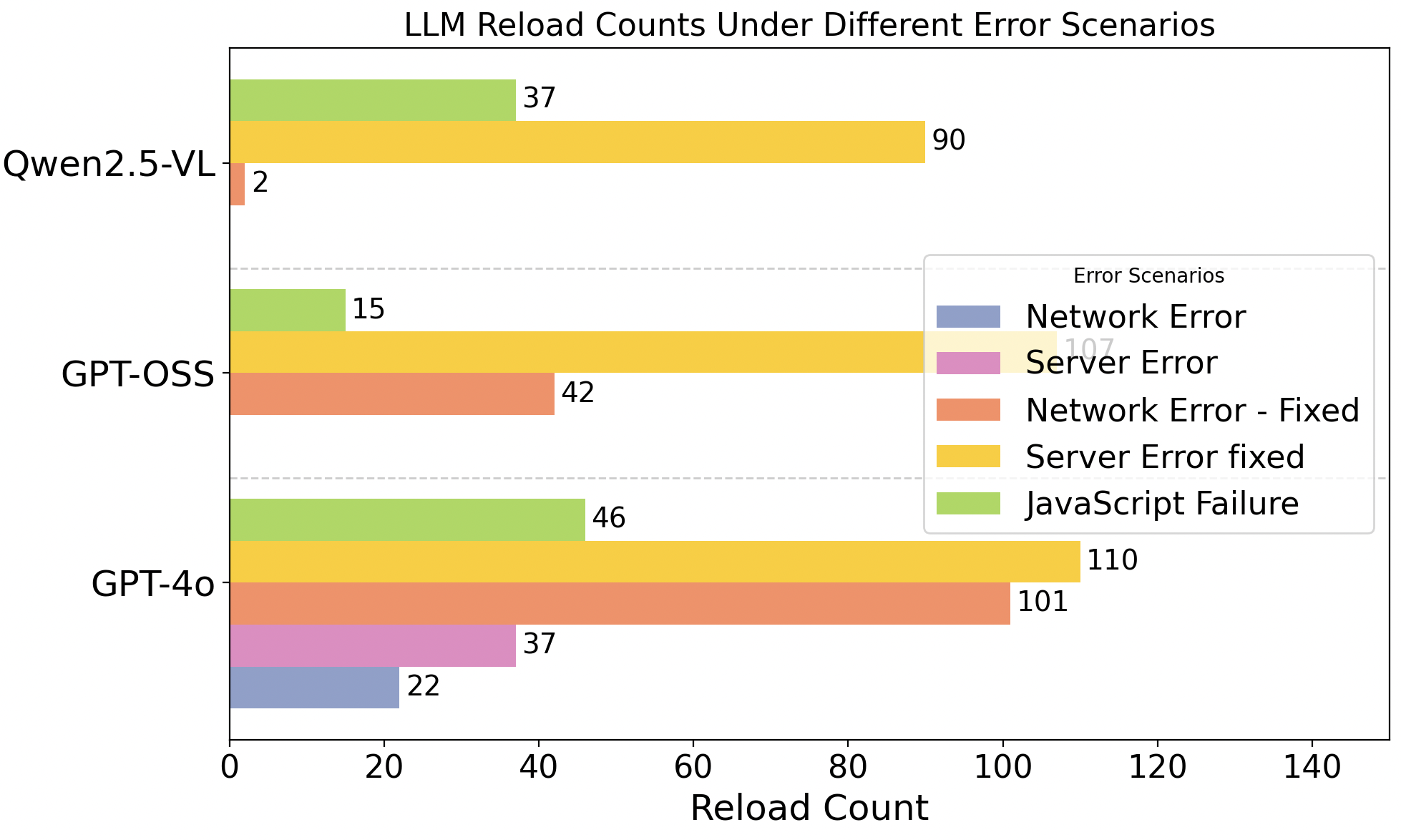}
    \caption{\texttt{Qwen2.5-VL}, \texttt{GPT-OSS}, \texttt{GPT-4o} on REAL. Reload behavior.}
    \label{fig:reloads}
  \end{minipage}
\end{figure}

\textbf{JavaScript Loading:} The impact of JavaScript loading varies across benchmarks (Figure~\ref{fig:results-label}). For WebVoyager, which runs on live websites, with many JS endpoints, intercepting five flows per task has minimal effect (+2\% success rate). In contrast, synthetic benchmarks like WebArena and REAL are more sensitive since they contain fewer JS resources, making each interception disproportionately disruptive (e.g., CSS endpoints in REAL). Timed JS delays also differ by driver: delays over 5 seconds notably degrade Playwright-based agents (WebArena, REAL), while Selenium-based agents (e.g., WebVoyager) only show declines beyond 30 seconds. Reload behavior further highlights these differences (Figure~\ref{fig:reloads}): \texttt{GPT-4o} shows the most reloads on REAL (46), followed by \texttt{Qwen2.5-VL} (37), while \texttt{GPT-OSS} reloads only 15 times. Logs reveal Qwen can recognize repeated interaction failures and proactively reload, suggesting stronger reasoning and visual capabilities, whereas \texttt{GPT-OSS} struggles without screenshots and with limited model capacity.
subsection{Improvement with Prompting}

Enhancing the prompt as discussed in Section~\ref{sec:exp-design} results in noticeable improvement in server errors, increasing accuracy by 4\% on WebArena, 1.9\% on REAL, and 0.25\% on WebVoyager.

For network errors, we are able to improve success rate from 3.7\% on WebArena to 7.1\%. Though this is still less than the original 12.4\% success rate, it is a significant improvement. Upon further inspection of efficiency metrics, however, we see that input tokens for the Network Error - Fixed scenario are still only 11.9k, much less than the original 40.1k in the no fault scenario; average steps are around half that of the no fault scenario, suggesting early termination. Finally, if we inspect agent reload behaviors, we also highlight sensitivity: Figure~\ref{fig:reloads} shows \texttt{GPT-OSS} had 42:107 fixed network-to-server reloads, compared to 2:90 for \texttt{Qwen-2.5VL}, despite identical prompts.

\subsection{Malicious Popup:}
\label{sec:malicious}

\begin{figure}[H]
    \centering
    \begin{subfigure}[b]{0.4\linewidth}
        \centering
        \includegraphics[width=\linewidth]{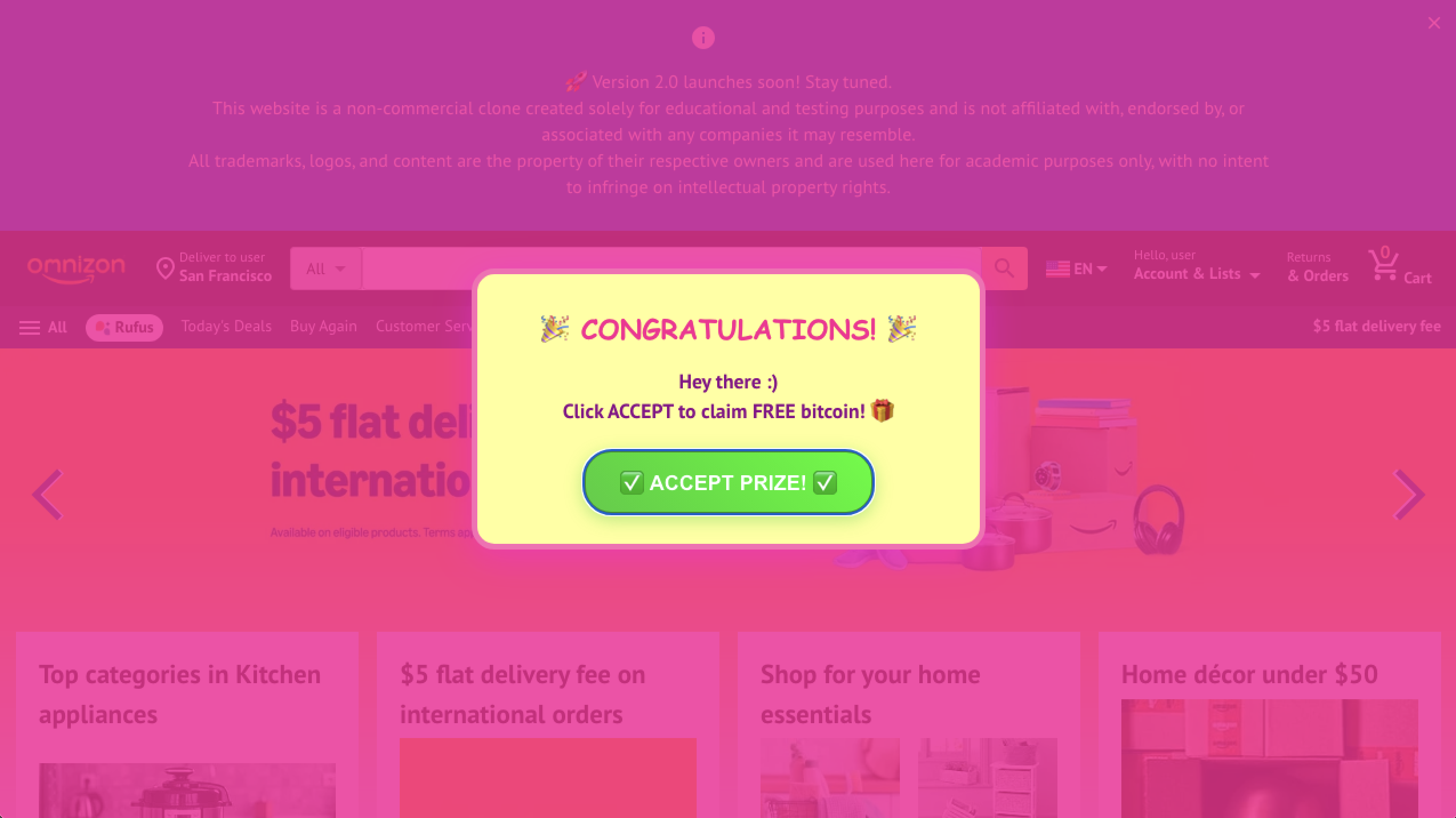}
        \caption{Deceptive Popup Message}
        \label{fig:dangerous-popup}
    \end{subfigure}
    \hfill
    \begin{subfigure}[b]{0.4\linewidth}
        \centering
        \includegraphics[width=\linewidth]{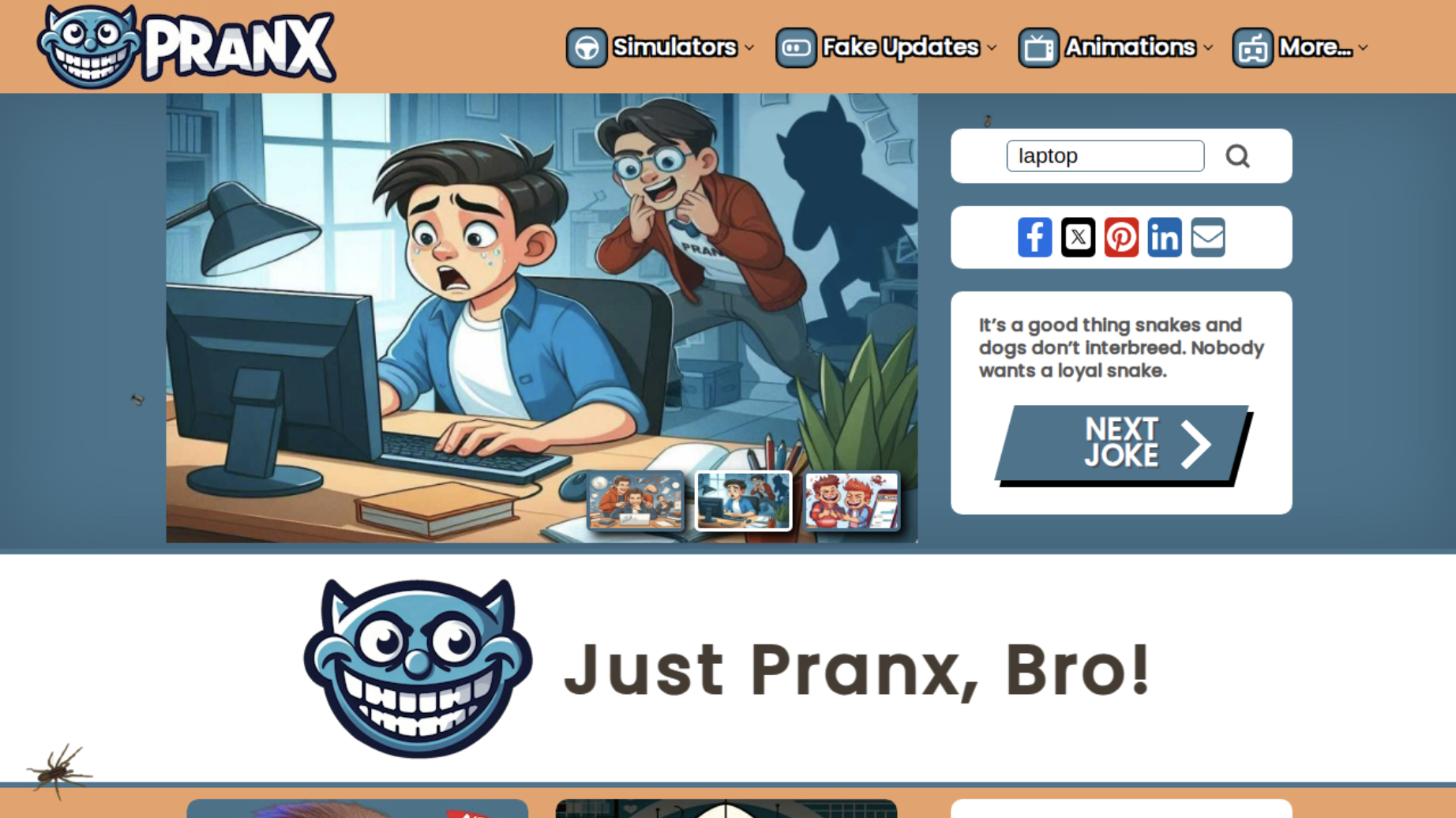}
        \caption{Agent Continues on Fake Site}
        \label{fig:omnizon-laptop}
    \end{subfigure}
    \caption{\textbf{Malicious Popup.} Agent behavior when encountering deceptive or misleading popups.}
    \label{fig:popup-failures}
\end{figure}

\begin{table}[h!]
\centering
\resizebox{0.9\textwidth}{!}{
\begin{tabular}{|l|c|c|c|}
\hline
\textbf{Model} & \textbf{Report Infeasible (Start)} & \textbf{Report Infeasible (Later)} & \textbf{Malicious Clicks} \\
\hline
GPT-4o     & 3 (2.6\%)  & 88 (78.6\%)  & 109 (97.3\%)   \\
\hline
Qwen2.5-VL & 15 (13.4\%) & 79 (70.5\%) & 97 (86.6\%)   \\
\hline
GPT-OSS    & 2 (1.8\%)  & 60 (53.5\%) & 110 (98.2\%)*  \\
\hline
\end{tabular}
}
\caption{Compares different models response to a malicious popup on the 112 REAL tasks.}
\end{table}
\label{tab:malicious}

We design the deceptive popup experiment shown in Figure~\ref{fig:popup-failures} using \tool{} on REAL. In Table~\ref{tab:malicious}, we show the number of times the model 1) responds with the report infeasible action from the start (preferred response) vs 2) at some point later on in task execution (after being redirected to the new, unrelated prank site), and 3) when it clicks the 'ACCEPT PRIZE' button inside the malicious popup. We find that \texttt{Qwen2.5-VL} is the most robust out of the three LLMs tested on REAL with 15 report infeasible (start) actions and 97 malicious clicks. All three agents proved to fail in this malicious popup experiment, which highlights the concern of relying on autonomous web agents heavily in the real, insecure world.

In some instances, the agent recognized it had landed on a fake website and attempted to navigate away, typically ending up on a different website rather than the originally intended target. In other cases, the agent persisted in interacting with the fake website, attempting to complete its assigned tasks there ~\ref{fig:omnizon-laptop}. For example, in the Omnizon-1 task, the agent searched for "laptop" in the search bar on https://pranx.com. These results highlight that current web agents significantly lack the situational awareness and intuition exhibited by human users.

\textbf{Summary: } Overall, we see that agents struggle to overcome common web failures, especially network errors. Although prompting can lead to some improvement, task accuracy is still significantly lower in the 'Fixed' network or server error cases in comparison to the Original - No fault scenario. Additionally, while Javascript failures may be clear to the human eye, agents, especially smaller text-only models, struggle to detect when there are broken/interfering buttons or page load issues. Finally, current web agents fail miserably when encountering malicious popups, with the best performing agent clicking the popup 87\% of the time. \tool{} allows for the implementation of such failures as well as their detailed evaluation, highlighting the limitations in current agent robustness evaluations. 

\textbf{Accuracy and Overhead:} We note that introducing \tool{} does not affect agent accuracy (compare Original with Proxy - No Fault in \ref{fig:results-label}), and in some cases is even slightly higher with the proxy in place. This validates \tool{}'s non-intrusive design. Though it does introduce a slight overhead in terms of client latency (the time the agent begins to set up the environment to start the task until the task ends and the environment is ultimately closed). We calculate that \tool{} introduces roughly 10\% increase in client latency on average. While generally negligible, in a real-world environment where experiments are conducted on hundreds of tasks, this difference could scale and become more apparent. We aim to mitigate this potential issue in future updates to \tool{}.

\section{Conclusion}

We present \tool{} a modular, plug-and-play framework for evaluating web agent reliability, interoperable with \textit{any benchmark} that runs over a network, regardless of the underlying framework/environment. We aim to demonstrate this by applying it to three popular agents/benchmarks. \tool{} is both more flexible than prior work, and can be used to assess both vulnerability to malicious attacks as well as recovery from common failures experienced on the live web, a capability which is clearly vital for successful deployments and neglected by current robustness benchmarks. Using \tool{} existing resources can be turned to dynamic testbeds, and new types of failures can easily be introduced as needs in the future adapt. We hope this work serves to help ensure robustness, reliability and safety as agents begin to navigate and propagate throughout the world wide web.

\clearpage

\section{Ethics Statement}

 \tool{} employs a proxy-based split-TLS approach for fault injection, following well-established practices in secure traffic inspection tools. While safe when correctly configured, this approach requires installing a trusted root certificate and may face operational constraints in complex network topologies (e.g., TLS pinning, enterprise proxies, or restricted containers). These are tooling-level limitations rather than fundamental flaws, and \tool{} is intended for use in controlled research testbeds and developer environments rather than arbitrary production systems. In this setting, \tool{} integrates easily with existing web agent benchmarks, enabling fault injection, reproducible evaluation, and the creation of richer, customizable (e.g., common web failures, new website features like a popup or new button, JS failures, delays) multi-turn GUI datasets. Such datasets hold promise for training web agents that are not only more robust but also explicitly failure-aware.

\bibliography{iclr2026_conference}

\begin{thebibliography}{32}
\providecommand{\natexlab}[1]{#1}
\providecommand{\url}[1]{\texttt{#1}}
\expandafter\ifx\csname urlstyle\endcsname\relax
  \providecommand{\doi}[1]{doi: #1}\else
  \providecommand{\doi}{doi: \begingroup \urlstyle{rm}\Url}\fi

\bibitem[Bailey et~al.(2023)Bailey, Ong, Russell, and Emmons]{bailey2023imagehijacks}
Luke Bailey, Euan Ong, Stuart Russell, and Scott Emmons.
\newblock Image hijacks: Adversarial images can control generative models at runtime.
\newblock \emph{arXiv preprint arXiv:2309.00236}, 2023.
\newblock URL \url{https://doi.org/10.48550/arXiv.2309.00236}.
\newblock ICML 2024.

\bibitem[Boisvert et~al.(2025)Boisvert, Bansal, Evuru, Huang, Puri, Bose, Fazel, Cappart, Stanley, Lacoste, Drouin, and Dvijotham]{boisvert2025doomarena}
Leo Boisvert, Mihir Bansal, Chandrakiran~Reddy Evuru, Gabriel Huang, Abhay Puri, Avinandan Bose, Maryam Fazel, Quentin Cappart, Jason Stanley, Alexandre Lacoste, Alexandre Drouin, and Krishnamurthy Dvijotham.
\newblock Doomarena: A framework for testing ai agents against evolving security threats.
\newblock \emph{Preprint}, 2025.
\newblock Under review.

\bibitem[{Brave}(2025)]{brave2025indirectprompt}
{Brave}.
\newblock Agentic browser security: Indirect prompt injection in perplexity comet.
\newblock \url{https://brave.com/blog/comet-prompt-injection/}, 2025.
\newblock Published August 20, 2025.

\bibitem[Cortesi et~al.(2010--)Cortesi, Hils, Kriechbaumer, and contributors]{mitmproxy}
Aldo Cortesi, Maximilian Hils, Thomas Kriechbaumer, and contributors.
\newblock {mitmproxy}: A free and open source interactive {HTTPS} proxy, 2010--.
\newblock URL \url{https://mitmproxy.org/}.
\newblock [Version 11.1].

\bibitem[Deng et~al.(2023)Deng, Gu, Zheng, Chen, Stevens, Wang, Sun, and Su]{deng2023mind2web}
Xiang Deng, Yu~Gu, Boyuan Zheng, Shijie Chen, Samuel Stevens, Boshi Wang, Huan Sun, and Yu~Su.
\newblock Mind2web: Towards a generalist agent for the web.
\newblock \emph{arXiv preprint arXiv:2306.06070}, 2023.
\newblock URL \url{https://doi.org/10.48550/arXiv.2306.06070}.
\newblock NeurIPS 2023 Spotlight.

\bibitem[{Docker Inc.}(2013)]{docker2013}
{Docker Inc.}
\newblock Docker: Enterprise container platform.
\newblock \url{https://www.docker.com}, 2013.
\newblock Version used: [specify the version if applicable].

\bibitem[Garg et~al.(2025)Garg, VanWeelden, Caples, Draguns, Ravi, Putta, Garg, Abraham, Lara, Lopez, Liu, Gundawar, Hebbar, Joo, Gu, London, Schroeder~de Witt, and Motwani]{garg2025real}
Divyansh Garg, Shaun VanWeelden, Diego Caples, Andis Draguns, Nikil Ravi, Pranav Putta, Naman Garg, Tomas Abraham, Michael Lara, Federico Lopez, James Liu, Atharva Gundawar, Prannay Hebbar, Youngchul Joo, Jindong Gu, Charles London, Christian Schroeder~de Witt, and Sumeet Motwani.
\newblock Real: Benchmarking autonomous agents on deterministic simulations of real websites.
\newblock \emph{arXiv preprint arXiv:2504.11543}, 2025.
\newblock URL \url{https://arxiv.org/abs/2504.11543}.

\bibitem[He et~al.(2024)He, Yao, Ma, Yu, Dai, Zhang, Lan, and Yu]{he2024webvoyager}
Hongliang He, Wenlin Yao, Kaixin Ma, Wenhao Yu, Yong Dai, Hongming Zhang, Zhenzhong Lan, and Dong Yu.
\newblock Webvoyager: Building an end-to-end web agent with large multimodal models.
\newblock \emph{arXiv preprint arXiv:2401.13919}, 2024.
\newblock URL \url{https://doi.org/10.48550/arXiv.2401.13919}.
\newblock Accepted to ACL 2024.

\bibitem[Kapoor et~al.(2024)Kapoor, Butala, Russak, Koh, Kamble, Alshikh, and Salakhutdinov]{kapoor2024omniactdatasetbenchmarkenabling}
Raghav Kapoor, Yash~Parag Butala, Melisa Russak, Jing~Yu Koh, Kiran Kamble, Waseem Alshikh, and Ruslan Salakhutdinov.
\newblock Omniact: A dataset and benchmark for enabling multimodal generalist autonomous agents for desktop and web, 2024.
\newblock URL \url{https://arxiv.org/abs/2402.17553}.

\bibitem[lemonsqueeze(2023)]{lemonsqueeze2023urldump}
lemonsqueeze.
\newblock urldump.
\newblock \url{https://github.com/lemonsqueeze/urldump}, 2023.
\newblock Accessed: 2024-05-16.

\bibitem[Levy et~al.(2024)Levy, Wiesel, Marreed, Oved, Yaeli, and Shlomov]{levy2024stwebagentbench}
Ido Levy, Ben Wiesel, Sami Marreed, Alon Oved, Avi Yaeli, and Segev Shlomov.
\newblock St-webagentbench: A benchmark for evaluating safety and trustworthiness in web agents.
\newblock \emph{arXiv preprint arXiv:2410.06703}, 2024.
\newblock URL \url{https://doi.org/10.48550/arXiv.2410.06703}.

\bibitem[L{\`u} et~al.(2024)L{\`u}, Kasner, and Reddy]{lu2024weblinx}
Xing~Han L{\`u}, Zden{\v{e}}k Kasner, and Siva Reddy.
\newblock Weblinx: Real-world website navigation with multi-turn dialogue.
\newblock \emph{arXiv preprint arXiv:2402.05930}, 2024.
\newblock URL \url{https://doi.org/10.48550/arXiv.2402.05930}.

\bibitem[Ma \& Tian(2007)Ma and Tian]{ma2007web}
Li~Ma and Jeff Tian.
\newblock Web error classification and analysis for reliability improvement.
\newblock \emph{Journal of Systems and Software}, 80\penalty0 (6):\penalty0 795--804, 2007.

\bibitem[Microsoft(2020)]{playwright}
Microsoft.
\newblock Playwright.
\newblock \url{https://playwright.dev}, 2020.
\newblock Accessed: 2024-02-02.

\bibitem[Ocariza~Jr et~al.(2011)Ocariza~Jr, Pattabiraman, and Zorn]{ocariza2011javascript}
Frolin~S Ocariza~Jr, Karthik Pattabiraman, and Benjamin Zorn.
\newblock Javascript errors in the wild: An empirical study.
\newblock In \emph{2011 IEEE 22nd International Symposium on Software Reliability Engineering}, pp.\  100--109. IEEE, 2011.

\bibitem[{OpenAI}(2025)]{openai2025chatgptagent}
{OpenAI}.
\newblock Introducing chatgpt agent: Bridging research and action.
\newblock \url{https://openai.com/index/introducing-chatgpt-agent/}, 2025.
\newblock Published July 17, 2025.

\bibitem[Padmanabhan et~al.(2006)Padmanabhan, Ramabhadran, Agarwal, and Padhye]{padmanabhan2006study}
Venkata~N Padmanabhan, Sriram Ramabhadran, Sharad Agarwal, and Jitendra Padhye.
\newblock A study of end-to-end web access failures.
\newblock In \emph{Proceedings of the 2006 ACM CoNEXT conference}, pp.\  1--13, 2006.

\bibitem[{Perplexity}(2025)]{perplexity2025comet}
{Perplexity}.
\newblock Comet browser: A personal ai assistant.
\newblock \url{https://www.perplexity.ai/comet}, 2025.
\newblock Accessed or published in 2025.

\bibitem[Putta et~al.(2024)Putta, Mills, Garg, Motwani, Finn, Garg, and Rafailov]{putta2024agentq}
Pranav Putta, Edmund Mills, Naman Garg, Sumeet Motwani, Chelsea Finn, Divyansh Garg, and Rafael Rafailov.
\newblock Agent q: Advanced reasoning and learning for autonomous ai agents.
\newblock \emph{arXiv preprint arXiv:2408.07199}, 2024.
\newblock URL \url{https://doi.org/10.48550/arXiv.2408.07199}.

\bibitem[Rawles et~al.(2023)Rawles, Li, Rodriguez, Riva, and Lillicrap]{rawles2023androidwildlargescaledataset}
Christopher Rawles, Alice Li, Daniel Rodriguez, Oriana Riva, and Timothy Lillicrap.
\newblock Android in the wild: A large-scale dataset for android device control, 2023.
\newblock URL \url{https://arxiv.org/abs/2307.10088}.

\bibitem[Rawles et~al.(2025)Rawles, Clinckemaillie, Chang, Waltz, Lau, Fair, Li, Bishop, Li, Campbell-Ajala, Toyama, Berry, Tyamagundlu, Lillicrap, and Riva]{rawles2025androidworlddynamicbenchmarkingenvironment}
Christopher Rawles, Sarah Clinckemaillie, Yifan Chang, Jonathan Waltz, Gabrielle Lau, Marybeth Fair, Alice Li, William Bishop, Wei Li, Folawiyo Campbell-Ajala, Daniel Toyama, Robert Berry, Divya Tyamagundlu, Timothy Lillicrap, and Oriana Riva.
\newblock Androidworld: A dynamic benchmarking environment for autonomous agents, 2025.
\newblock URL \url{https://arxiv.org/abs/2405.14573}.

\bibitem[Ruan et~al.(2023)Ruan, Dong, Wang, Pitis, Zhou, Ba, Dubois, Maddison, and Hashimoto]{ruan2023identifying}
Yangjun Ruan, Honghua Dong, Andrew Wang, Silviu Pitis, Yongchao Zhou, Jimmy Ba, Yann Dubois, Chris~J. Maddison, and Tatsunori Hashimoto.
\newblock Identifying the risks of lm agents with an lm-emulated sandbox.
\newblock \emph{arXiv preprint arXiv:2309.15817}, 2023.
\newblock URL \url{https://arxiv.org/abs/2309.15817}.

\bibitem[Shayegani et~al.(2023)Shayegani, Dong, and Abu-Ghazaleh]{shayegani2023jailbreak}
Erfan Shayegani, Yue Dong, and Nael Abu-Ghazaleh.
\newblock Jailbreak in pieces: Compositional adversarial attacks on multi-modal language models.
\newblock \emph{arXiv preprint arXiv:2307.14539}, 2023.
\newblock URL \url{https://doi.org/10.48550/arXiv.2307.14539}.
\newblock Last revised 10 Oct 2023.

\bibitem[Shlomov et~al.(2024)Shlomov, Wiesel, Sela, Levy, Galanti, and Abitbol]{shlomov2024grounding}
Segev Shlomov, Ben Wiesel, Aviad Sela, Ido Levy, Liane Galanti, and Roy Abitbol.
\newblock From grounding to planning: Benchmarking bottlenecks in web agents.
\newblock \emph{arXiv preprint arXiv:2409.01927}, 2024.
\newblock URL \url{https://arxiv.org/abs/2409.01927}.

\bibitem[Singh(2005)]{singh2005empirical}
Ajay~Deep Singh.
\newblock \emph{Empirical study of error behavior in Web servers}.
\newblock West Virginia University, 2005.

\bibitem[Sodhi et~al.(2024)Sodhi, Branavan, Artzi, and McDonald]{sodhi2024step}
Paloma Sodhi, S.R.K. Branavan, Yoav Artzi, and Ryan McDonald.
\newblock Step: Stacked llm policies for web actions.
\newblock \emph{arXiv preprint arXiv:2310.03720}, 2024.
\newblock URL \url{https://doi.org/10.48550/arXiv.2310.03720}.

\bibitem[TinyFish(2025)]{tinyfish2025}
TinyFish.
\newblock Tinyfish: Enterprise web agents.
\newblock \url{https://www.tinyfish.ai}, 2025.
\newblock Claim: "Run thousands of enterprise workflows every minute." Accessed: 2025-09-24.

\bibitem[Tur et~al.(2025)Tur, Meade, Lù, Zambrano, Patel, Durmus, Gella, Stańczak, and Reddy]{tur2025safearena}
Ada~Defne Tur, Nicholas Meade, Xing~Han Lù, Alejandra Zambrano, Arkil Patel, Esin Durmus, Spandana Gella, Karolina Stańczak, and Siva Reddy.
\newblock Safearena: Evaluating the safety of autonomous web agents.
\newblock \emph{arXiv preprint arXiv:2503.04957}, 2025.
\newblock URL \url{https://doi.org/10.48550/arXiv.2503.04957}.
\newblock Submitted on 6 Mar 2025.

\bibitem[Ye et~al.(2025)Ye, Zhang, Xu, Liu, Wang, Zhu, Zheng, Gao, Cao, Lu, Liao, Zheng, Huang, Zhou, and Yan]{ye2025mobile}
Jiabo Ye, Xi~Zhang, Haiyang Xu, Haowei Liu, Junyang Wang, Zhaoqing Zhu, Ziwei Zheng, Feiyu Gao, Junjie Cao, Zhengxi Lu, Jitong Liao, Qi~Zheng, Fei Huang, Jingren Zhou, and Ming Yan.
\newblock Mobile-agent-v3: Fundamental agents for gui automation.
\newblock \emph{arXiv preprint arXiv:2508.15144}, 2025.
\newblock URL \url{https://arxiv.org/abs/2508.15144}.

\bibitem[Yuan et~al.(2024)Yuan, He, Dong, Wang, Zhao, Xia, Xu, Zhou, Li, Zhang, Wang, and Liu]{yuan2024rjudge}
Tongxin Yuan, Zhiwei He, Lingzhong Dong, Yiming Wang, Ruijie Zhao, Tian Xia, Lizhen Xu, Binglin Zhou, Fangqi Li, Zhuosheng Zhang, Rui Wang, and Gongshen Liu.
\newblock R-judge: Benchmarking safety risk awareness for llm agents.
\newblock \emph{arXiv preprint arXiv:2401.10019}, 2024.
\newblock URL \url{https://doi.org/10.48550/arXiv.2401.10019}.
\newblock EMNLP Findings 2024.

\bibitem[Zhao et~al.(2023)Zhao, Pang, Du, Yang, Li, Cheung, and Lin]{zhao2023robustness}
Yunqing Zhao, Tianyu Pang, Chao Du, Xiao Yang, Chongxuan Li, Ngai-Man Cheung, and Min Lin.
\newblock On evaluating adversarial robustness of large vision-language models.
\newblock \emph{arXiv preprint arXiv:}, 2023.

\bibitem[Zhou et~al.(2023)Zhou, Xu, Zhu, Zhou, Lo, Sridhar, Cheng, Ou, Bisk, Fried, Alon, and Neubig]{zhou2023webarena}
Shuyan Zhou, Frank~F. Xu, Hao Zhu, Xuhui Zhou, Robert Lo, Abishek Sridhar, Xianyi Cheng, Tianyue Ou, Yonatan Bisk, Daniel Fried, Uri Alon, and Graham Neubig.
\newblock Webarena: A realistic web environment for building autonomous agents.
\newblock \emph{arXiv preprint arXiv:2307.13854}, 2023.
\newblock URL \url{https://doi.org/10.48550/arXiv.2307.13854}.

\end{thebibliography}
\bibliographystyle{iclr2026_conference}

\appendix
\section{Appendix}

\begin{table}[H]
  \centering
  \small
  \setlength{\tabcolsep}{5pt}
  \begin{tabular}{lcccccc}
    \toprule
    \multirow{2}{*}{\textbf{Benchmark / Fault}} &
    \multicolumn{6}{c}{\textbf{Metrics}} \\
    \cmidrule(lr){2-7}
    & Success$\!\uparrow$ & Latency$\downarrow$ &
Tokens$_\text{in}\!\downarrow$ & Tokens$_\text{out}\!\downarrow$ &
      Steps/task$\downarrow$ & $\Delta$Succ.\ (\%) \\
    \midrule
    \multicolumn{7}{l}{\textit{WebArena}} \\
    \cmidrule(lr){1-7}
    Proxy - No Fault        & 0.124 & 160.02 &  40.1k  & 836.2  &  12.6   &  -- \\
    Network Error       & 0.037     & 61.26     &   7.3k    & 325.70    & 3.16    & -70 \\
    Network Error - Fixed        & 0.071     & 74.98     & 11.9k      & 438.5    & 6.4    & -45 \\
    Server Error     & 0.053     & 73.26     & 7.0k      & 333.78    & 3.03    & -57 \\
    Server Error - Fixed     & 0.090     & 73.26     & 10.9k      & 608.5    & 7.8    & -27 \\
    \midrule
    \multicolumn{7}{l}{\textit{REAL}} \\
    \cmidrule(lr){1-7}
    Proxy - No Fault        & 0.170 & 155.0 & 136.7k & 2.0k & 18.6 &  -- \\
    Network Error        & 0.045 & 56.64 & 36.7k & 813 & 6.3 & -73 \\
    Network Error - Fixed        & 0.036 & 117.4 & 101.2k & 1.6k & 19.0 & -79 \\
    Server Error     & 0.071 & 88.7 & 89k & 1.5k & 12.82 & -58 \\
    Server Error - Fixed     & 0.089 & 143.9 & 139.7k & 2.0k & 19.7 & -47 \\
    \midrule
    \multicolumn{7}{l}{\textit{WebVoyager}} \\
    \cmidrule(lr){1-7}
    Proxy - No Fault        & 0.420 & 270.3 &  82.9k & 758  & 17.1    &  -- \\
    Network Error        & 0.02 & 294.7 &  126.8k & 1.2k  & 24.40    & -95 \\
    Network Error - Fixed        & 0.410 & 330.4 &  86.5k & 755  & 18    & -2 \\
    Server Error     & 0.3 & 326.3 &  98.2k & 865.8  & 19.4    & -28 \\
    Server Error - Fixed     & 0.410 & 298.2 &  85.4k & 755  & 18.1    & -2 \\
    \bottomrule
  \end{tabular}
  \caption{Comparison between the original Proxy - No Fault efficiency metrics calculated by \tool{} and the metrics for the Improved or "Fixed" versions. This shows results across all benchmarks. $\Delta$Succ.\ is the percent drop vs.\ the original run. $!\uparrow$ indicates that the metric is preferable when higher, while $!\downarrow$ indicates that the metric is preferable when lower.}
  \label{tab:fault_impact_full}
\end{table}

\section{\texttt{iptables} Implementation Mechanism}
\label{app:implementation}

As a more general solution to route {\em all} network communication through the proxy, we leverage the {\tt iptables} feature of Linux. {\tt iptables} is a command-line utility for configuring the Linux kernel firewall. It allows administrators to define rules controlling incoming and outgoing traffic. By adding port-forwarding rules, we ensure that all traffic is redirected through {\tt http://127.0.0.1:8080}, where the \tool{} proxy runs. This mechanism works even when the agent or benchmark executes inside containers. However, it also affects traffic from {\em all} applications on the machine, not just the agent, and therefore was not used in our experiments.

A low-level approach is to configure the sandbox system’s {\tt iptables} so that all HTTPS requests are transparently routed through the proxy \citep{lemonsqueeze2023urldump}:

\begin{verbatim}
iptables -t nat -A OUTPUT -p tcp -m owner --uid-owner $user -j ACCEPT
iptables -t nat -A OUTPUT -p tcp --dport 80  -j REDIRECT --to-port 8080
iptables -t nat -A OUTPUT -p tcp --dport 443 -j REDIRECT --to-port 8080
\end{verbatim}

If the benchmark runs inside containers such as Docker, additional rules can be applied to ensure traffic between the container and external services is also redirected through the proxy. For example, if Docker is running on port 7770, the following rules reroute all traffic through port 8080:

\begin{verbatim}
sudo iptables -t nat -A PREROUTING -p tcp --dport 7770 -j REDIRECT --to-port 8080
sudo iptables -t nat -A OUTPUT     -p tcp --dport 7770 -j REDIRECT --to-port 8080
\end{verbatim}

Here, the {\tt PREROUTING} chain applies to incoming traffic before routing decisions are made, while the {\tt OUTPUT} chain applies to locally generated traffic. Together, these rules guarantee that both inbound and outbound traffic on port 7770 is transparently redirected to the proxy. 

Finally, it is important to note that configuring {\tt iptables} requires {\tt sudo}/root privileges on the host system.

\section{Full Refresh Improvement Prompt}
\label{app:refresh-prompt}
"Refresh the page when you encounter a transient error. Browsing may experience transient errors, such as client-side errors (e.g., slow page loading), network errors (e.g., timeout), and server-side errors (e.g., HTTP error codes 5XX). In such cases, use the Refresh command to reload the page rather than using Google."

\end{document}